\documentclass[10pt,twocolumn,letterpaper]{article}

\usepackage{cvpr}              

\usepackage[dvipsnames]{xcolor}

\usepackage{ulem}

\newcommand{\cmmntWord}[1]{}

\definecolor{cvprblue}{rgb}{0.21,0.49,0.74}
\usepackage[pagebackref,breaklinks,colorlinks,citecolor=cvprblue]{hyperref}
\usepackage[accsupp]{axessibility} 

\def\paperID{14490} 
\def\confName{CVPR}
\def\confYear{2024}

\title{Turb-Seg-Res: A Segment-then-Restore Pipeline for Dynamic Videos with Atmospheric Turbulence}

\author{
Ripon Kumar Saha\textsuperscript{1}, \hspace{1em}
Dehao Qin\textsuperscript{2}, \hspace{1em}
Nianyi Li\textsuperscript{2}, \hspace{1em}
Jinwei Ye\textsuperscript{3}, \hspace{1em}
Suren Jayasuriya\textsuperscript{1}\\[1ex] 
{ \textsuperscript{1}Arizona State University,} 
{ \textsuperscript{2}Clemson University,} 
{ \textsuperscript{3}George Mason University} \\
{\small \texttt{\{rsaha8, sjayasur\}@asu.edu, \{dehaoq, nianyil\}@clemson.edu, jinweiye@gmu.edu}}}

\begin{document}
\maketitle
\begin{abstract}
Tackling image degradation due to atmospheric turbulence, particularly in dynamic environment, remains a challenge for long-range imaging systems. Existing techniques have been primarily designed for static scenes or scenes with small motion. This paper presents the first segment-then-restore pipeline for restoring the videos of dynamic scenes in turbulent environment. We leverage mean optical flow with an unsupervised motion segmentation method to separate dynamic and static scene components prior to restoration. After camera shake compensation and segmentation, we introduce foreground/background enhancement leveraging the statistics of turbulence strength and a transformer model trained on a novel noise-based procedural turbulence generator for fast dataset augmentation. Benchmarked against existing restoration methods, our approach restores most of the geometric distortion and enhances sharpness for videos. We make our code, simulator, and data publicly available to advance the field of video restoration from turbulence: \href{https://riponcs.github.io/TurbSegRes/}{riponcs.github.io/TurbSegRes}
\end{abstract}    
\section{Introduction}
\label{sec:intro}

Atmospheric turbulence poses a considerable impediment to high-fidelity long-range imaging, inducing geometric distortions and blur that severely compromise image quality~\cite{roggemann1996imaging, chen2014detecting, crittenden1978effects}. This issue becomes particularly acute in long-range horizontal and slant path imaging in environments with extreme temperature gradients. Such unpredictable variations in air density and velocity induce characteristic distortions including random pixel displacements, referred to as tilt~\cite{chan2022tilt}, blur, and contrast reduction. This presents a formidable challenge for current vision algorithms to consistently detect, classify, and delineate objects within such degraded images.

Conventional multi-frame techniques such as ``lucky frame" or geometric stabilization~\cite{TCIMultiPurdue2020, MultiPCA2007atmospheric, MultiRPCAStabilization2019, MultiKmeanBspine2016, MultiRegsiDeblur2014, MultiDeep2022} perform well in static or slightly dynamic scenes but struggle in highly dynamic environments with significant scene or camera motion. Supervised deep learning-based methods~\cite{Turbnet2022, AtNet_2021learning} leverage turbulence simulators to generate training data~\cite{Simulator0_chimitt2020simulating,mao2021accelerating}. While these models show potential, they often exhibit temporal artifacts when applied to real-world dynamic video sequences with moving objects.

The limitation of previous methods for dynamic video stems from the compounded effect of object motion and atmospheric turbulence. These factors disrupt optical flow maps which makes image registration difficult for multi-frame techniques. In this paper, we address this challenge by proposing a unique segment-then-restore pipeline. This method utilizes integrated optical flows computed between frames for unsupervised motion segmentation, which then enables the individual processing of static background and dynamic foreground elements. We use a weighted image stacking technique for background enhancement, and a transformer model for both foreground and background sharpening trained on data from a novel tilt-and-blur turbulence simulator. Our approach notably reduces background distortions while maintaining the clarity of foreground details throughout the sequence.

Our specific contributions include:

\begin{itemize}
\item An unsupervised segmentation method that utilizes integrated optical flow with optimized number of frames calculated to segment a frame into static background and moving foreground. 
\item An adaptive Gaussian-weighted image stacking method for background processing which utilizes physics-based turbulence strength statistics for optimal frame selection across a range of turbulence intensities.
\item An image restoration transformer model trained on simulated data generated from a novel turbulence video simulator that incorporates both tilt and adaptive blur based on procedural noise to generate plausible turbulence effects. 
\end{itemize}
To validate our approach, we conduct experiments on two well-established datasets, CLEAR~\cite{anantrasirichai2013atmospheric},  OTIS~\cite{OTIS2017}. We also augment the recent URG-T~\cite{dehaoUnsupervised2023} segmentation dataset with additional real videos captured in the wild that feature extreme turbulence conditions. We compare to state-of-the-art baselines for video restoration in this domain. Finally, we release the code, simulator, and data for our study: \href{https://riponcs.github.io/TurbSegRes}{riponcs.github.io/TurbSegRes}

\section{Related Work}
\label{sec:related work}

\textbf{Modeling and Simulating Turbulence:} The physics of turbulence was first detailed by Kolmogorov~\cite{kolmogorov,kolmogorov1991local}, and we refer the reader to several reference books~\cite{ishimaru1999wave,tatarski2016wave} including imaging through turbulence~\cite{roggemann1996imaging,kopeika1998system}. Ihrke et al.~\cite{ihrke2007eikonal} and Gutierrez et al.~\cite{gutierrez2006simulation} helped advance the understanding and simulation of refractive and atmospheric phenomena, respectively. 

In addition to modeling, there has been concerted effort to simulate the effects of turbulence. Potvin et al.~\cite{potvin2007parametric} presented a parametric model for simulating turbulence effects on imaging systems. Similarly, Repasi and Weiss~\cite{repasi2011computer} proposed a computer simulation of image degradations by atmospheric turbulence for horizontal views. Reinhardt et al.~\cite{reinhardt2016efficient} developed computationally efficient techniques to simulate optical atmospheric refraction phenomena. 

Instead of simulating 3D random turbulence fields, Schwartzman et al.~\cite{schwartzman2017turbulence} directly created 2D random physics-based distortion vector fields. A series of simulators that utilize Zernike coefficents have been developed~\cite{Simulator0_chimitt2020simulating, mao2021accelerating} culminating in a real-time phase-to-space simulation~\cite{9969142}. In our work, we do not utilize these physics-based simulators because they are targeted for single image creation, but develop a procedural noise-based turbulence video simulator for coherent turbulence effects.


\textbf{``Lucky Image" Reconstruction Methods:} Fried~\cite{fried1978probability} introduced the concept of ``lucky imaging" for astronomical imaging through turbulence, and these frames are typically short-exposure to minimize blur~\cite{bensimon1981measurement}. Lucky fusion combines sharp image patches to form the final diffraction limited image using a variety of methods~\cite{aubailly2009automated,anantrasirichai2013atmospheric,zhu2012removing,xie2016removing}. However, these techniques require a static scene with no motion for lucky imaging to be effective.

Contrasting prior work, Mao et al.~\cite{TCIMultiPurdue2020} pioneered restoration for dynamic sequences leveraging optical-flow guided lucky imaging coupled with blind deconvolution. We avoid lucky image techniques in favor of a segment-then-restore framework for dynamic scene restoration.


\indent \textbf{Deep Learning for Turbulence Restoration:} With the help of the synthetic datasets, several supervised neural networks have recently been proposed to address turbulence restoration. A complex-valued convolutional neural network was proposed~\cite{ANANTRASIRICHAI202369} for reducing geometric distortions and subsequent refinement of micro-details. AT-Net utilized epistemic uncertainty in their network design for single frame reconstruction~\cite{AtNet_2021learning}. Mao et al.~\cite{Turbnet2022} introduced TurbNet, a physics-inspired transformer model for single frame restoration. Zhang et al.~\cite{zhang2024imaging} also proposed an efficient transformer-based network but for video sequences. 

In addition, there have been unsupervised models proposed in the literature. Li et al.~\cite{Li_2021_ICCV} utilize implicit neural representations (INRs) to estimate grid deformation for single image restoration. Jiang et al.~\cite{jiang2023nert} extended this INR approach to have a more realistic tilt-and-blur model, but still targeted to static scenes.

\indent \textbf{Segmentation in Atmospheric Turbulence:} Segmentation in atmospheric turbulence is relatively understudied in the literature. Cui and Zhang~\cite{8901108} developed a supervised network for semantic segmentation in turbulence, trained on synthetic datasets. Recently Qin et al.~\cite{dehaoUnsupervised2023} proposed a two-stage unsupervised segmentation for the video affected by atmospheric turbulence, but is computationally expensive. Our segmentation, while conceptually simple, performs near state-of-the-art while having very low latency.

\section{Method}
Our method is driven by the observation that in dynamic scenes, moving objects and the static background experience distinct motion types: moving objects are influenced by their own motion plus turbulence and camera movements, while the static background is solely affected by turbulence and camera motions. We efficiently segment moving objects from the static background, even amidst turbulence, and apply targeted enhancement strategies to each.

\begin{figure*}
    \centering
    \includegraphics[width=17.4cm]{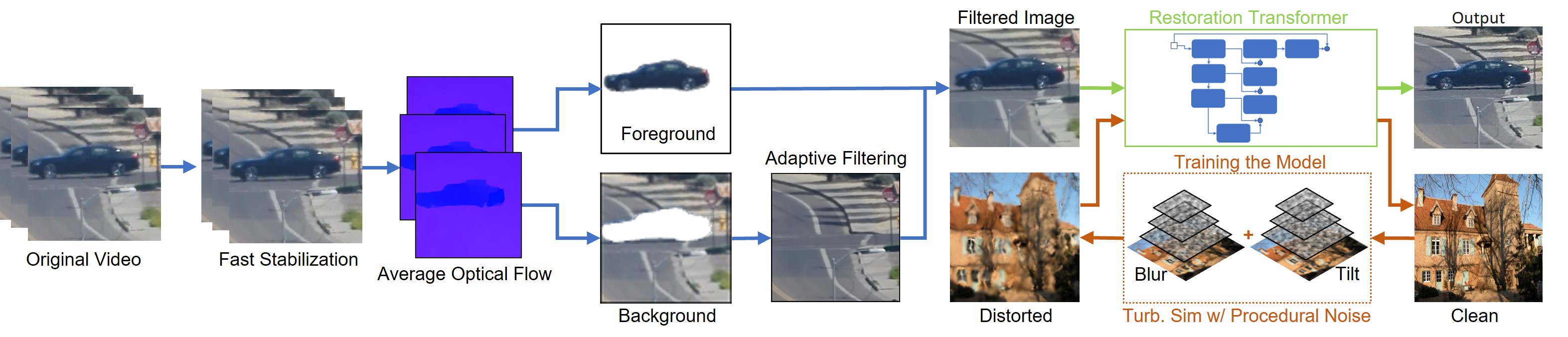}
    \caption{Our pipeline processes dynamic video frames by first stabilizing image sequences using normalized cross-correlation, followed by segmenting moving objects using average optical flow. The background is processed with adaptive filtering and blended with a separately extracted foreground. The background and segmented foreground are seamlessly merged using Poisson pyramid blending. Finally, a transformer architecture, trained on our simulator, refines the combined images. }
    \label{fig:architecture}
\end{figure*}

In a nutshell, our full method, visualized in ~\cref{fig:architecture} performs the following steps: (1) video stabilization to reduce camera motion; (2) motion segmentation into dynamic foreground regions and static background; (3) weighted image stacking for the static background; (4) Poisson blending to merge the enhanced background with the dynamic foreground regions; and finally (5) sharpening the entire image using a trained transformer model.

\subsection{Stabilization Methodology}
Long range imaging with high focal length cameras are extremely sensitive to vibrations which leads to camera motion. Most existing video stabilization methods~\cite{videostabilzationsurvey} rely on local feature detection/matching (e.g. SIFT, SURF, and ORB) prone to error due to turbulence-induced distortions. We instead implement a simple, yet effective method of estimating global camera motion via cross-correlation and applying the estimated translation to stabilize the frame. 

Let \(\mathbf{V} = \{\mathbf{V}_i\}\) be the sequence of video frames, each normalized to \([0, 1]\). We perform the following steps: (1) subtract the mean intensity of the entire video sequence from each frame: \( \mathbf{V}'_i = \mathbf{V}_i - \mu \); (2) select the first frame \( \mathbf{F}_{\text{ref}} = \mathbf{V}'_1 \); (3) crop each frame as \(\mathbf{V}_{\text{crop}}^{(i)}\) with a border of 50 pixels (representing the maximum camera shift possible) to avoid edge effects; (4) calculate the cross-correlation ($\text{xcorr2d}$) of the cropped frame and the reference frame (implemented as a convolution in practice without flipping the kernel), and extract the coordinate positions of the maximum correlation:  \begin{equation}\mathbf{H}^{(i)} = \text{xcorr2d}\left(\mathbf{F}_{\text{ref}}, \mathbf{V}_{\text{crop}}^{(i)}\right)\end{equation}
\begin{equation}
(\Delta x_i, \Delta y_i) = \arg \max(\mathbf{H}^{(i)}) - \left(\frac{H}{2}, \frac{W}{2}\right) 
\end{equation}
With the extracted $(\Delta x_i, \Delta y_i)$, we translate the selected frame via a matrix multiplication in homogeneous space to perform video stabilization.


This stabilization method, both simple and effective as demonstrated in the supplemental material, gains efficiency through our GPU-accelerated convolution approach, improving upon the CPU-based \texttt{scikit-image match\_template} function in SciPy. This adaptation achieves a $1000\times$ speedup in the convolution phase, significantly enhancing processing speed for high-resolution images. Tested on videos with camera motion between 0 to 110 pixels and on $1080p$ synthetic video with max 400 pixels vibration, it ensured error-free vibration compensation.

\subsection{Motion Segmentation in Turbulent Conditions}
We introduce an automated motion segmentation technique that effectively discriminates between turbulence-induced distortions and inherent object motion for dynamic scenes. Our approach accounts for the Gaussian distribution nature~\cite{fried1965statistics, fried1982anisoplanatism} of turbulence effects, which often makes capturing object movements challenging~\cite{chan2022tilt, dehaoUnsupervised2023}.

The cornerstone of our method is a multi-frame optical flow analysis based on pre-trained RAFT~\cite{teed2020raft}. Conventional frame-by-frame optical flow struggles in the face of turbulence. Our technique computes the average optical flow (AOF) across a dynamically determined number of neighboring frames, formulated as follows:
\begin{equation}
    \text{AOF} = \frac{1}{N} \sum_{i=1}^{N} ||OF(F_{\text{current}}, F_{i})||,
\end{equation}
where \( F_{\text{current}} \) is the current frame, \( F_{i} \) denotes the \( i \)-th neighboring frame, and $||OF(\cdot,\cdot)||$ is the magnitude of the optical flow. The value of \( N\) is adaptively selected to enhance the segmentation clarity between static and dynamic regions. This optimization is achieved by maximizing the average distance of the normalized optical flow values from a median value of 0.5, ensuring a clear separation between static (near 0) and dynamic (near 1) elements:
\begin{equation}
N_{\text{opt}} = \arg\max_{N} \left( \frac{1}{M} \sum_{i=1}^{M} \left| \text{AOF}(N)_i - 0.5 \right| \right),
\end{equation}
where \( M \) is the total number of pixels in the optical flow mask, and \(\text{AOF}(N)_i\), representing the optical flow magnitude value for the \( i \)-th pixel normalized to the range \([0, 1]\). This allows for an adaptive response to varying motion displacements within turbulent scenes. 

\begin{figure}[htbp]
    \centering
    \includegraphics[width=1\columnwidth]{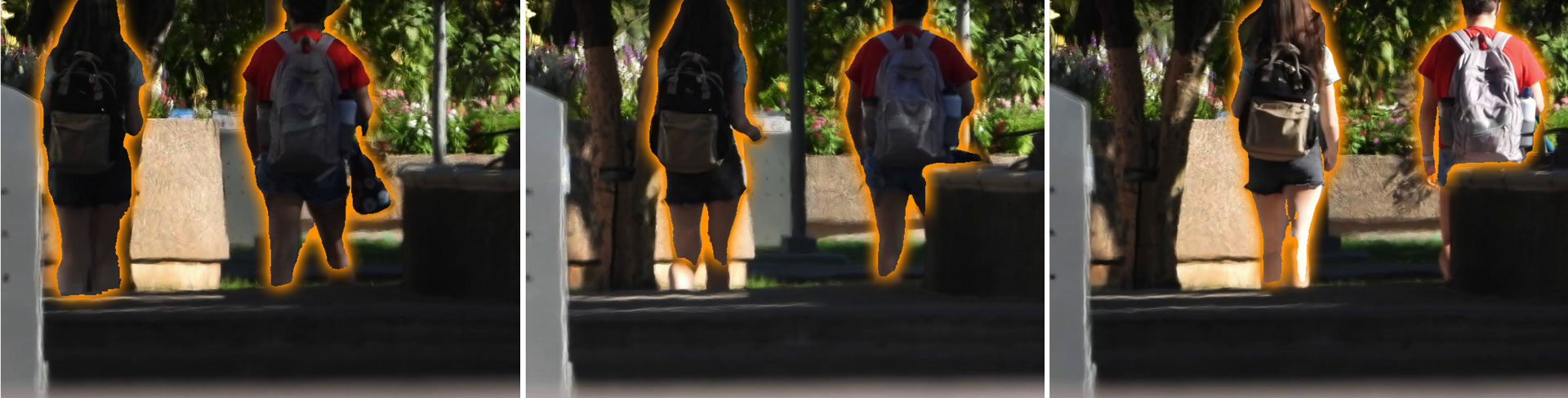}
    \caption{Visual representation of motion segmentation across consecutive frames in a video sequence. Our proposed method effectively separates moving subjects (students walking) visualized by the golden outlines.}
    \label{fig:segmentation_oursVisual}
\end{figure}

Finally, we threshold our calculated AOF to separate the video into static background and dynamic background. While this segmentation technique is simple (leveraging average optical flow), we find it highly effective in practice as shown in ~\cref{fig:segmentation_oursVisual}, which displays segmentation results across three distinct frames of a video clip. We also conducted ablation on the choice of segmentation in~\cref{sec:ablation} that justify this simpler method is competitive with state-of-the-art unsupervised methods~\cite{dehaoUnsupervised2023,cho2023treating,LIU2022103700,ye2022deformable} for turbulent video while having lower latency.

\subsection{Background Processing}

For static backgrounds, we implement tilt correction via weighted averaging, guided by atmospheric turbulence strength as indicated by \(C_n^2\)~\cite{saha2022turbulence} values derived from video analysis. This method adapts Gaussian weighted averaging relative to detected turbulence, targeting precise stabilization and geometric distortion reduction in the background. \(C_n^2\) can be calculated via image information via the following equation~\cite{saha2022turbulence, zamek2006turbulence}: 
\begin{equation}
    C_n^2 = \frac{PFOV^{2} \times D^{\frac{1}{3}}}{L\times P} \times \frac{\sigma (\mathbf{V})^{2}}{Grad(\mathbf{V})}, \label{eq:cn2}
\end{equation}
\noindent where \( \mathbf{V} \) is the sequence of images, \( PFOV \) is the pixel field of view, \( D \) is the lens aperture diameter, \( L \) is the distance to target, and \( P \) is the turbulence constant parameter, adjusted based on the scale of turbulence~\cite{zamek2006turbulence}. Lower \(C_n^2\) values, indicative of minimal turbulence, necessitate smaller Gaussian windows as the distortion is negligible. Higher \(C_n^2\) values signal stronger turbulence, requiring larger windows to average more frames for effective distortion correction. Simplified to \(C_n^2 \propto \frac{\sigma (\mathbf{V})^{2}}{Grad(\mathbf{V})}\), this adaptive process culminates in an averaged background with notably reduced geometric distortion. It efficiently addresses a spectrum of turbulence strengths, facilitating optimal correction automatically without needing manual hyperparameter adjustments.


\subsection{Video Restoration}
After segmentation and background enhancement, our pipeline's final step is restoration, where we merge the dynamic foreground into the background using Poisson blending~\cite{blending2023poisson}.  We train a transformer-based model to perform restoration for both the foreground and background simultaneously. This is the only supervised part of our machine learning pipeline, and thus requires a dataset of images without distortion paired with the corresponding turbulence-degraded images. Since real data is difficult to acquire such ground truth pairs, in the following subsections we describe our novel simulator based on simplex noise for rapidly generating data for our training. Then we proceed to describe the transformer model that is trained on this data. 

\subsubsection{Turbulence Video Simulator}
There have been existing turbulence-distortion simulators introduced in the prior literature~\cite{Simulator0_chimitt2020simulating, Simulator1_2011Optical, Simulator2_2011Simple, repasi2011computer, Simulator4_bos2012technique, Simulator5_leonard2012simulation, Simulator6_hardie2017simulation, schwartzman2017turbulence, Simulator9_2018synthesis, Simulator10_2019zernikecalc}. However, these simulators are targeted mostly for single image generation, and thus suffer from coherency or temporal artifacts when applied frame-by-frame for video. Recent hybrid or physics-based simulators show promise~\cite{Simulator0_chimitt2020simulating,mao2021accelerating}, however, we find their latency restrictive when generating varying levels of turbulence for each image for fast data creation\footnote{At the time of publication, a real-time physics-based simulator code implementation~\cite{9969142} was made available that could also potentially be used for training our transformer}. 

\textbf{Tilt Simulation:}
Our approach is tailored to model turbulence for image restoration purposes, focusing on creating plausible turbulence warp/tilt and blur effects rather than simulating turbulence with high physical accuracy. By employing 3D simplex noise, a procedural noise function from computer graphics for creating textures and volumetric effects~\cite{olano2002real}, we generate temporally coherent distortions. This method offers better scalability for high-dimensional noise and low latency and memory requirements, making it ideally suited for fast data creation for training models. The utilization of simplex noise reflects our focus on plausible video distortions over precise physical modeling of turbulence~\cite{lagae2010survey}:
\begin{equation}
s(x, y, t) = \sum_{i=0}^{N-1} 2^i A_i \cdot \text{snoise}(f_i \cdot x, f_i \cdot y, f_i \cdot t)
\label{eq:noise}
\end{equation}

\noindent Where $A_i$ is the amplitude that scales the influence of each octave and $f_i$ is the frequency that dictates the granularity of the noise pattern. Utilizing an \(N = 8\) octave simplex noise method in~\cref{eq:noise}, we generate two sets of 3D noise, \(\mathcal{N}_x(H, W, T)\) and \(\mathcal{N}_y(H, W, T)\) where $H$ is height, $W$ is weight, $T$ is time, to simulate turbulence in the X and Y directions, respectively. This structure ensures that each pixel \((x, y)\) in the input image is dynamically shifted according to these noise sets across frames, effectively mimicking atmospheric turbulence tilt over time. Employing the $\text{snoise function}$\footnote{We utilize the Github package \url{https://github.com/caseman/noise}} ensures spatial and temporal coherence. With frequency ranges (\(f_i\)) from 1.5\% to 6\%, our approach simulates a spectrum of turbulence effects from mild to intense, ensuring qualitative fidelity to advanced models with operational simplicity and efficiency.

\textbf{Blur Simulation:} 
Adopting a similar approach to tilts, we utilize a set of 3D Perlin noise~\cite{PerlinNoise} to impose atmospheric blurring on tilt-adjusted frames. Generating Perlin noise with parameters such as width, height, depth, base frequency, and a level count of 11 allows dynamic modulation of blur intensity. This ensures spatial and temporal coherence with tilt magnitudes, enhancing realism. Adaptive Gaussian blurring adjusts the blur (\(\sigma\)) based on the 3D Perlin noise map and tilt map values, effectively simulating the variability of atmospheric scattering. This results in frames exhibiting authentically-varied blur levels, contributing to the simulation’s overall authenticity.

\textbf{Simulation Results and Comparisons:}
Our tilt-and-blur video simulator, capable of processing $200\times200$ resolution videos at interactive rates ($>14$ fps) on a consumer desktop, sets a new standard for efficiency and flexibility in turbulence simulation. Unique in its ability to produce coherent time-dependent distortions, it generates frames in just 100ms without precomputation, enabling the rapid creation of 100K short clips at up to $1024\times1024$ resolution for transformer model training. This performance surpasses general-purpose simulators and demonstrates substantial advantages over methods like the PS2 simulator~\cite{Simulator0_chimitt2020simulating}, particularly in generating training data with realistic temporal dynamics and without extensive pre-computation. A sample result is illustrated in ~\cref{fig:simulationFig}, with qualitative comparisons to a physics-based simulator shown in supplementary materials.
\begin{figure}
  \centering
  \begin{subfigure}{0.32\columnwidth}
    \includegraphics[width=\linewidth]{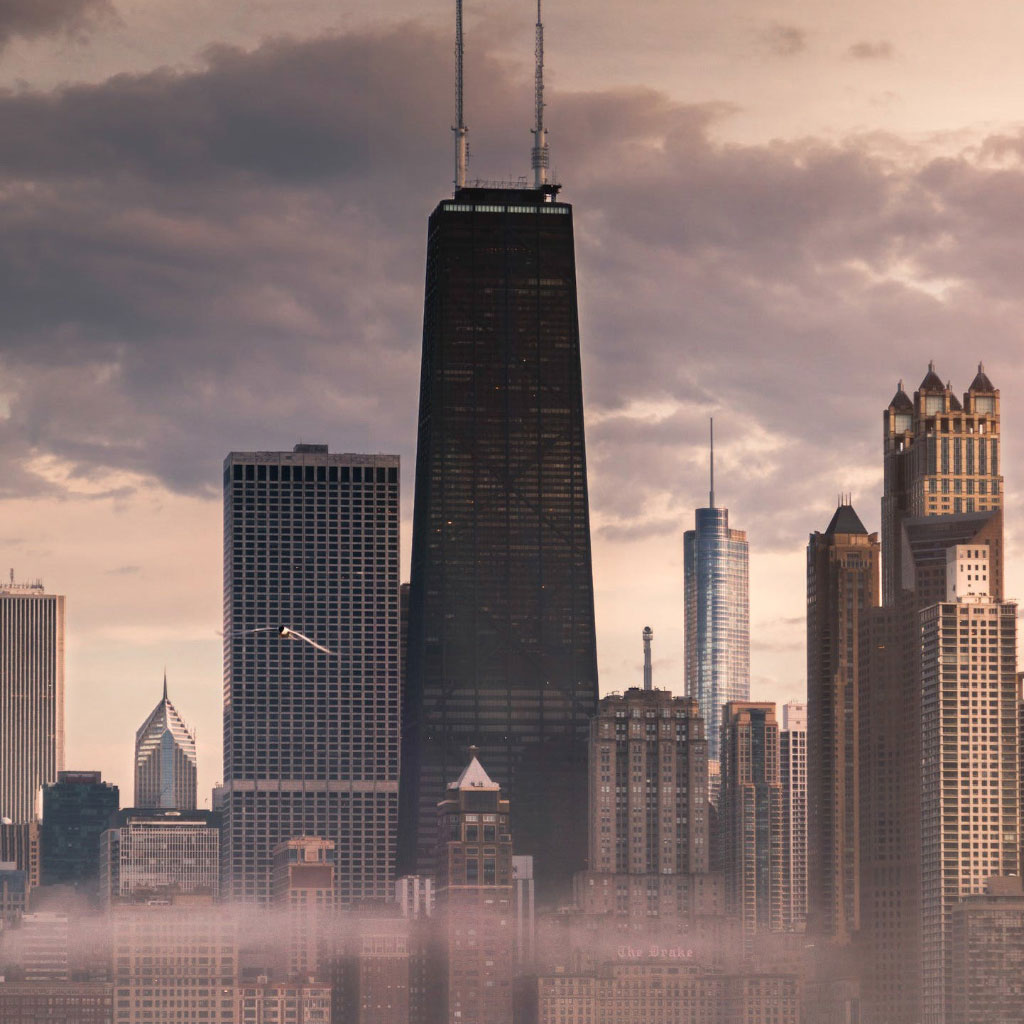}
    \caption{Original}
  \end{subfigure}
  \hfill 
  \begin{subfigure}{0.32\columnwidth}
    \includegraphics[width=\linewidth]{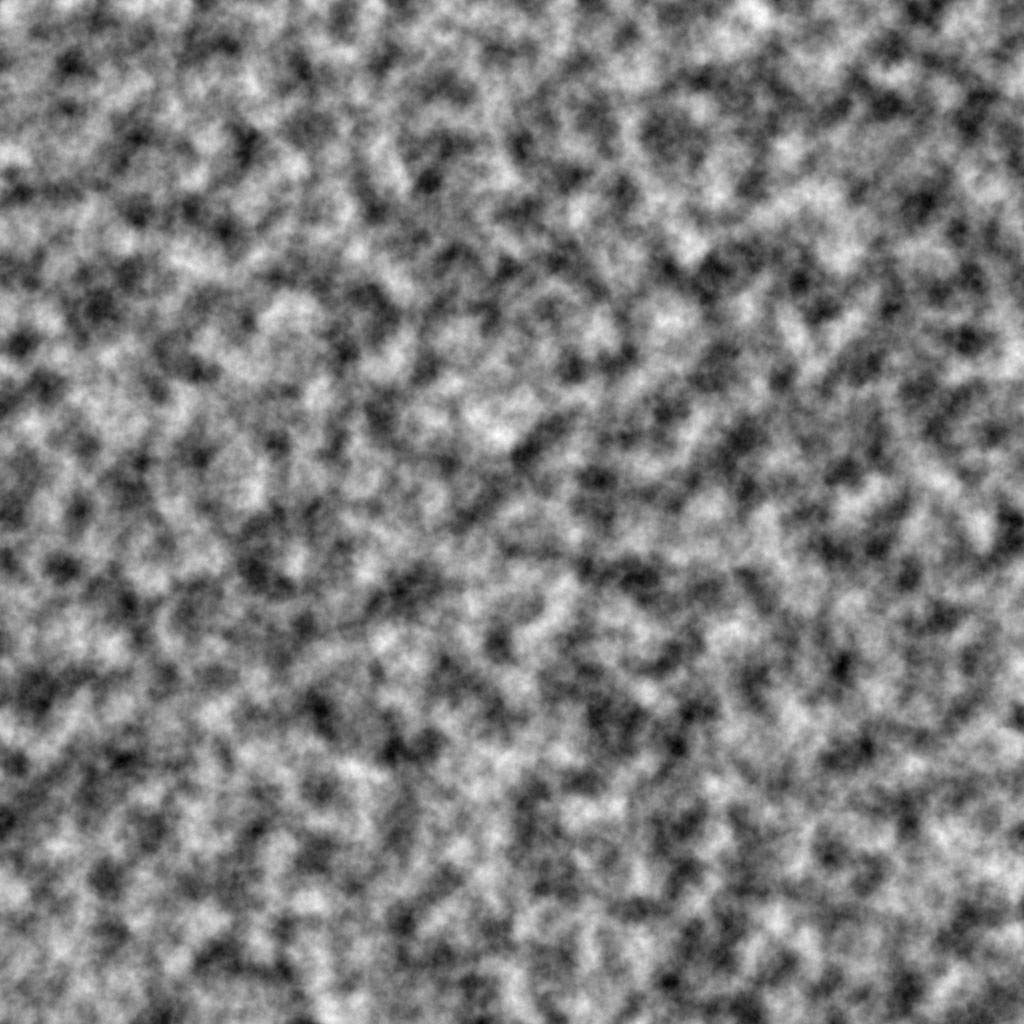}
    \caption{Noise}
  \end{subfigure}
  \hfill 
  \begin{subfigure}{0.32\columnwidth}
    \includegraphics[width=\linewidth]{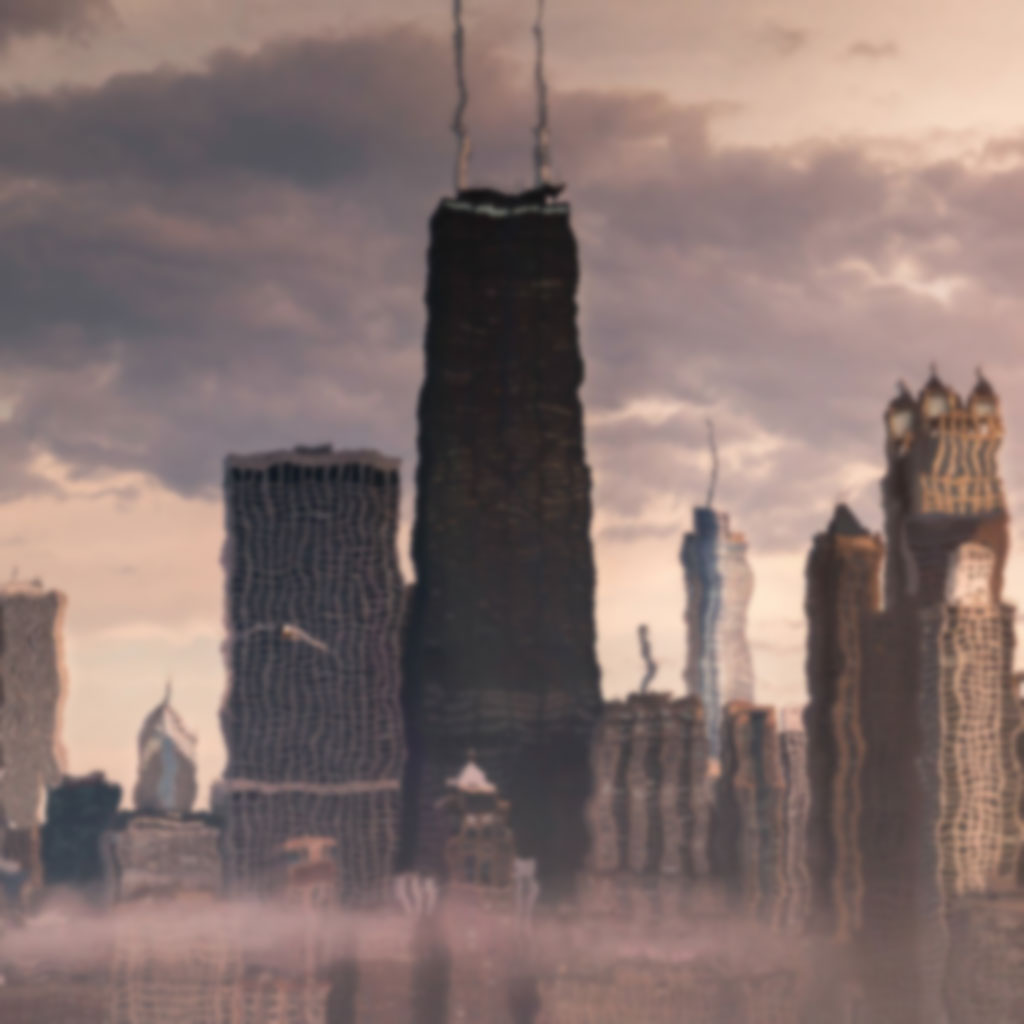}
    \caption{Turbulence}
  \end{subfigure}
  \caption{Comparison of the original image, noise-added image, and the image with simulated turbulence.}
  \label{fig:simulationFig}
\end{figure}
\subsubsection{Training the Transformer Model}

Our pipeline is compatible with any supervised deblurring model that can be trained from scratch on our procedurally generated, simulated turbulence data. For the final version of our pipeline, we leverage the image restoration transformer model, Restormer~\cite{zamir2022restormer}, due to its specific advantages in handling the complexities of turbulence-induced distortions. The Restormer model is particularly adept at dealing with the unpredictable and diverse nature of turbulence, thanks to its efficient attention mechanism that can focus on different regions of an HD image with varying distortion levels. Additionally, its capability to preserve and restore high-frequency details is crucial in maintaining the integrity of fine features in HD images, which are often adversely affected by turbulence. 

\section{Datasets and Implementation}

\subsection{Datasets}

\textbf{CLEAR Dataset~\cite{anantrasirichai2013atmospheric}:} The CLEAR dataset is a common benchmark for atmospheric turbulence mitigation~\cite{anantrasirichai2013atmospheric}. The data includes (1) eight sequences capturing a variety of objects 3.5m away with turbulence induced by a set of 8 gas stoves, and (2) three scenes captured outdoors at various long distances with actual atmospheric turbulence. The sequences were captured with a Canon EOS-1D Mark IV camera with 105mm lens for (1) and 400mm lens for (2).

\textbf{OTIS Dataset~\cite{OTIS2017}:} The OTIS dataset includes image sequences specifically designed for atmospheric turbulence mitigation studies. It features (1) controlled sequences with synthetic turbulence affecting various objects, generated through heat sources, and (2) natural scenes captured under real-world atmospheric turbulence. These images were captured with high-quality cameras, offering diverse scenarios to test turbulence mitigation techniques.

\textbf{Augmented URG-T Dataset~\cite{dehaoUnsupervised2023}:}
The URG-T dataset~\cite{dehaoUnsupervised2023} consists of 20 videos recorded in outdoor environments at long-range with corresponding ground truth masks for motion segmentation~\cite{dehaoUnsupervised2023}. Scenes include urban outdoor scenes including moving vehicles, airplanes, and pedestrians. Each clip features up to 56 frames at $1920\times1080$ resolution, shot with a Nikon Coolpix P1000, utilizing a 539mm focal length (effectively 3000mm due to sensor crop).

The URG-T dataset~\cite{dehaoUnsupervised2023} lacks scenes with static backgrounds, essential for testing our restoration pipeline. To compensate, we added a collection of static videos capturing various atmospheric turbulence levels, using a Nikon Coolpix P1000 with $125\times$ zoom. We filmed additional scenes in a desert under summer conditions, with subjects placed 100 meters to 1 kilometer away, at $1080p$ resolution.

\subsection{Training Details}

We run our method on an NVIDIA A100 GPU, using the AdamW optimizer with a 0.0003 initial learning rate for up to 60,000 iterations. Batch and patch sizes dynamically adjust, ranging from 28 to 4 and 128 to 384 pixels, respectively, following a predefined schedule. Training is completed approximately in one day.

\subsection{Comparison to State-of-the-Art}
We compare our method to the following state-of-the-art methods for turbulence restoration:
\begin{enumerate}
\item  AT-Net~\cite{AtNet_2021learning}: Deep learning framework that employs epistemic uncertainty analysis for effective restoration of images affected by atmospheric turbulence. 
\item TCI 2020~\cite{TCIMultiPurdue2020}: Physics-based models that integrates space-time non-local averaging for enhanced turbulence mitigation in both static and dynamic sequences. 
\item TurbNet~\cite{Turbnet2022}: Transformer-based model designed for atmospheric turbulence imaging, adept at extracting dynamic distortion maps and restoring turbulence-free images.
\end{enumerate}


\subsection{Latency}
Our final pipeline's latency per frame for each stage is listed in ~\cref{tab:per_frame_latency} for 100\%, 50\%, and 25\% of $1080p$ resolution. The main bottleneck is the calculation of the AOF and corresponding segmentation, due to pre-trained RAFT~\cite{teed2020raft} and the forward pass through the sharpening transformer. For competing methods, the latency per frame for $1080p$ resolution on a NVIDIA RTX 3090 GPU is: TCI - 4200s; AT-Net - 80s; Turbnet - 5s, while our pipeline latency per frame is 5.71s on a NVIDIA A100 GPU.
\begin{table}[ht!]
  \centering
  \caption{Per frame latency for operations at different resolutions}
  \label{tab:per_frame_latency}
  \begin{tabular}{lccc}
    \toprule
    Operations & \multicolumn{3}{c}{Size (pixels)} \\
    \cmidrule(lr){2-4}
    & 1920x1080 & 960x540 & 480x270 \\
    \midrule
    Read/Convert & 0.08s & 0.08s & 0.08s \\
    Vibration Calc. & 0.03s & 0.014s & 0.006s \\
    Stabilize & 0.12s & 0.05s & 0.03s \\
    Segmentation & 1.06s & 1.06s & 1.06s \\
    Gaussian Mean & 0.34s & 0.09s & 0.02s \\
    Combine BG/FG & 0.38s & 0.11s & 0.026s \\
    Sharpening & 3.7s & 1.0s & 0.25s \\
    \midrule
    \textbf{Total} & 5.71s & 2.404s & 1.472s \\
    \bottomrule
  \end{tabular}
\end{table}
\subsection{Quantitative Metrics}
For evaluating performance on the CLEAR dataset with available turbulence-free images, we employ standard metrics such as PSNR and SSIM, alongside IoU metrics to assess our pipeline's segmentation efficacy against URG-T's benchmarks~\cite{dehaoUnsupervised2023}.

\textbf{Line Deviation Metric:} For real-world scenarios lacking ground truth, we introduce a novel metric focusing on a system's capacity to preserve the integrity of straight lines during image restoration under turbulence. This metric, rooted in the observation that accurate restoration should correct distortions affecting straight structural elements (e.g., building edges), leverages Canny edge detection and the Probabilistic Hough Line Transform~\cite{Matas2000RobustDO} to quantify angular deviations from true vertical/horizontal orientations. A reduced mean deviation score per image indicates superior performance in maintaining geometric fidelity in turbulent conditions, with our method's effectiveness quantified through rolling mean and standard deviation across sequences.
\section{Experimental Results}

\subsection{Qualitative Comparison}
\begin{figure*}[ht!]
    \centering
    \includegraphics[width=17.4cm]{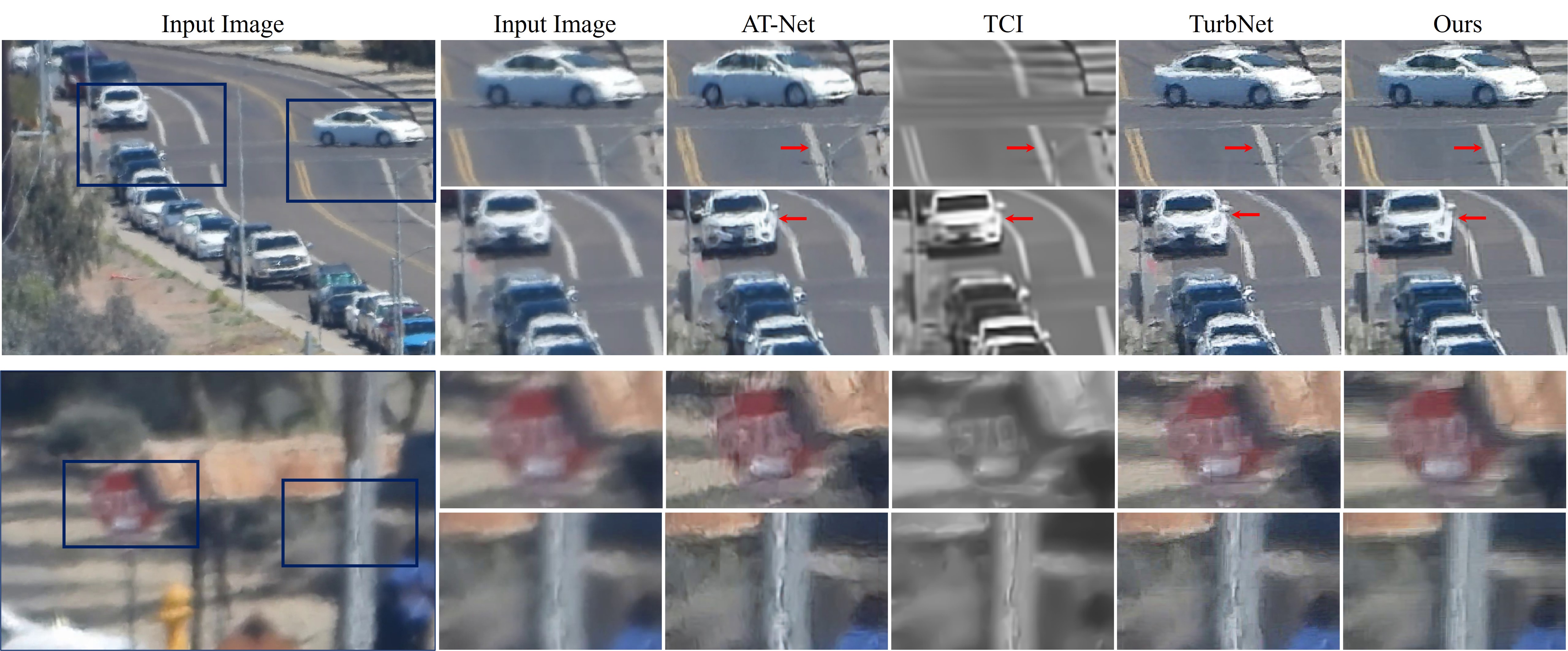}
    \caption{Comparative visualization of turbulence mitigation methods: AT-Net and Turbnet introduce artifacts, failing to clearly define edges of the road, while TCI maintains edge integrity but blurs moving objects. Our method shows minimal distortion with clear depiction of both static and dynamic elements.}
    \label{fig:GrandPicture}
\end{figure*}

\textbf{Visual Results on Real Turbulence Video:} In ~\cref{fig:GrandPicture}, we showcase zoomed in regions from single frames of two video sequences captured as part of the URG-T dataset. AT-Net and Turbnet are single-frame deep learning approaches that, while effective in some respects, introduce a significant number of artifacts into the processed images (see exaggerated sharpening effects on the edges of objects). TCI\footnote{Note that TCI reconstructs in grayscale per the original paper~\cite{TCIMultiPurdue2020}. We also developed a color extension for the TCI method by running on color channels independently, which we show in the supplemental material.}, is a multi-frame approach that excels at removing distortion but struggles with moving objects, e.g. the missing moving car in the first region or distortion in the second car.

Our method, in contrast, maintains the integrity of several object edges, presenting them solidly with minimal distortion, and importantly, it also accurately renders moving objects like cars, preserving their visibility and detail. \textit{We recommend viewing the supplemental video to closely evaluate our restoration performance across scenes.}

\begin{figure}[htbp]
    \centering
    \includegraphics[width=1\columnwidth]{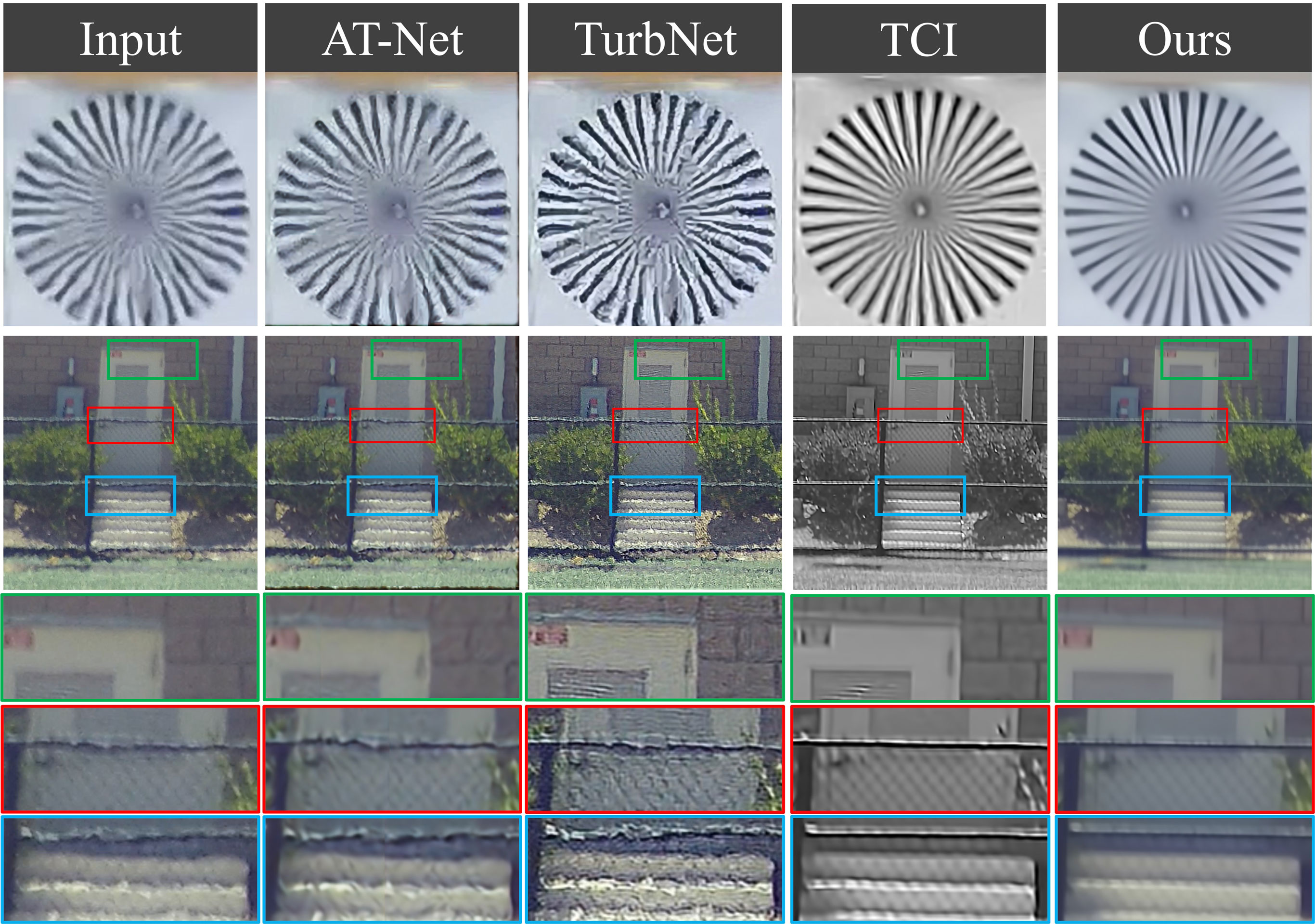}
    \caption{Analysis of the OTIS dataset reveals distinct outcomes. The upper row, featuring `Pattern16' (300 frames, $135\times135$ pixels), demonstrates notable clarity in our method compared to AT-Net and TurbNet, which exhibit significant distortion in finer details. The lower row, showcasing `Fixed Background (Door)' (300 frames, $520\times520$ pixels), highlights our approach's superiority in maintaining geometric integrity, especially in fence structures and the brickwork of stairs and doors. While TCI performs well on the fence, it distorts the circular patterns. Our method excels in preserving straight lines and enhancing edge definition in circular patterns, offering the clearest and least distorted results.}
    \label{fig:OTIS}
\end{figure}

\textbf{OTIS Reconstruction Results:} In ~\cref{fig:OTIS}, we showcase two scenes from the OTIS dataset (each video containing 300 frames): (1) a star chart target imaged under turbulence, and (2) a house entrance with a fence partially occluding it. The input images exhibit significant distortion, yet our method manages to produce images with the sharpest edges and without artifacts. For instance, in the Door image, the staircases appear crooked in results from other methods, and the foliage is indiscernible behind the fence. In contrast, our method leverages all 300 frames to correct the alignment of the staircase, and a closer examination shows we uniquely render the fence visible across the images with a continuous straight line.

\subsection{Quantitative \& Qualitative Comparison}
\begin{figure}[htbp]
    \centering
    \includegraphics[width=\columnwidth]{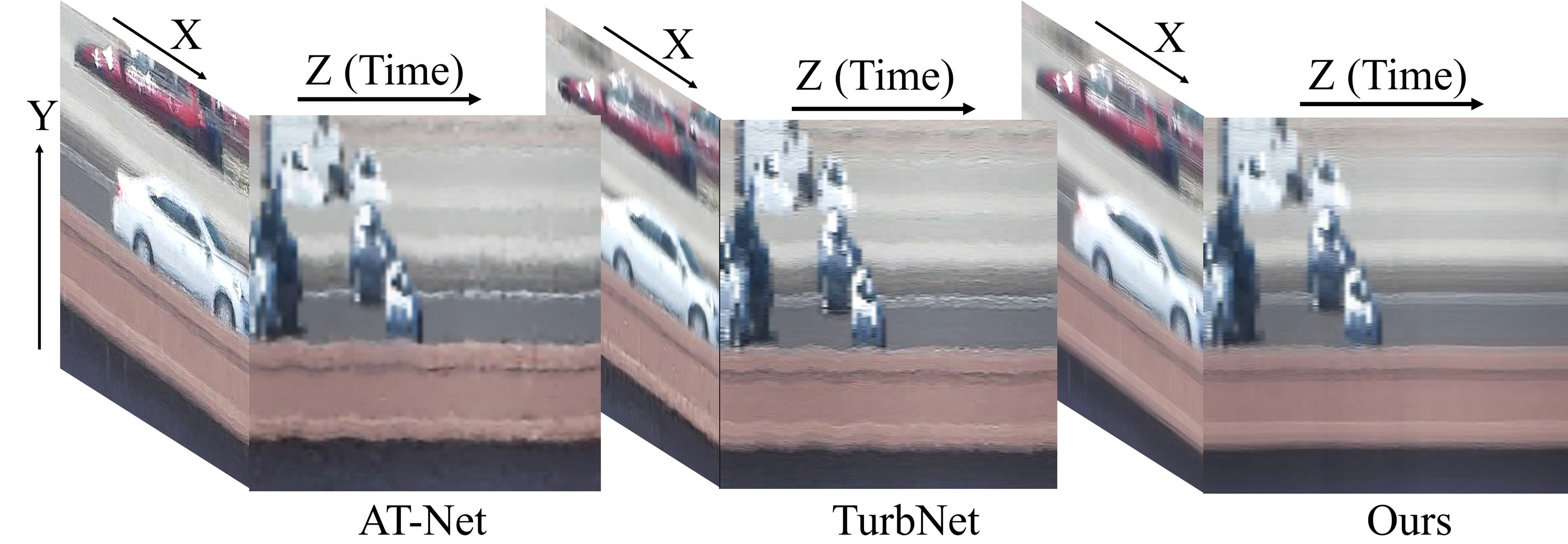}
    \caption{Demonstration of the Turb-Seg-Res method's stability in the time domain, highlighting its effectiveness in reducing pixel fluctuations in turbulent conditions.}
    \label{fig:TemporalTextureMap}
\end{figure}

\textbf{Temporal Consistency Comparison:} Our Turb-Seg-Res method demonstrates remarkable stability in the time domain, effectively mitigating the common issue of pixel fluctuation or the ``image dancing" effect typically observed in turbulent conditions. This stability is evident when visually comparing video sequences, as our approach maintains consistent pixel behavior across frames as shown in ~\cref{fig:TemporalTextureMap}.

\begin{table}[ht]
\centering
\caption{In the CLEAR dataset evaluation, images smaller than $400\times400$ pixels are excluded due to the fixed size requirements of AT-Net ($256\times256$) and Turbnet ($400\times400$). Larger images were processed using a patch-based approach. When we included all images of clear dataset, we got an average PSNR of 28.09 and SSIM of 0.935.}
\resizebox{\columnwidth}{!}{%
\begin{tabular}{c|ccc|ccc}
\hline
\textbf{Turb.} & \multicolumn{3}{c|}{\textbf{PSNR (dB)}} & \multicolumn{3}{c}{\textbf{SSIM}} \\
                     & \textbf{AT-Net} & \textbf{Turbnet} & \textbf{Ours} & \textbf{AT-Net} & \textbf{Turbnet} & \textbf{Ours} \\ \hline
Low                    & 22.91          & 18.86            & \textbf{26.48}         & 0.737          & 0.572            & \textbf{0.853}         \\
Medium                    & 23.37          & 18.99            & \textbf{26.50}        & 0.770          & 0.589            & \textbf{0.867}         \\
High                    & 22.62          & 19.15            & \textbf{24.82}         & 0.759          & 0.588            & \textbf{0.828}         \\ \hline
\end{tabular}
}
\label{tab:clear_dataset_results}
\end{table}

\textbf{CLEAR Dataset Comparison:} We ran quantitative evaluation of our method on the CLEAR dataset in ~\cref{tab:clear_dataset_results}. As one can see, our method achieved superior PSNR and SSIM values compared to competing methods, showing the benefit of our background processing and sharpening transform. Note that we do not need to perform segmentation as there is no dynamic scene motion in the CLEAR dataset. 

\begin{figure}[htbp]
    \centering
    \includegraphics[width=1\columnwidth]{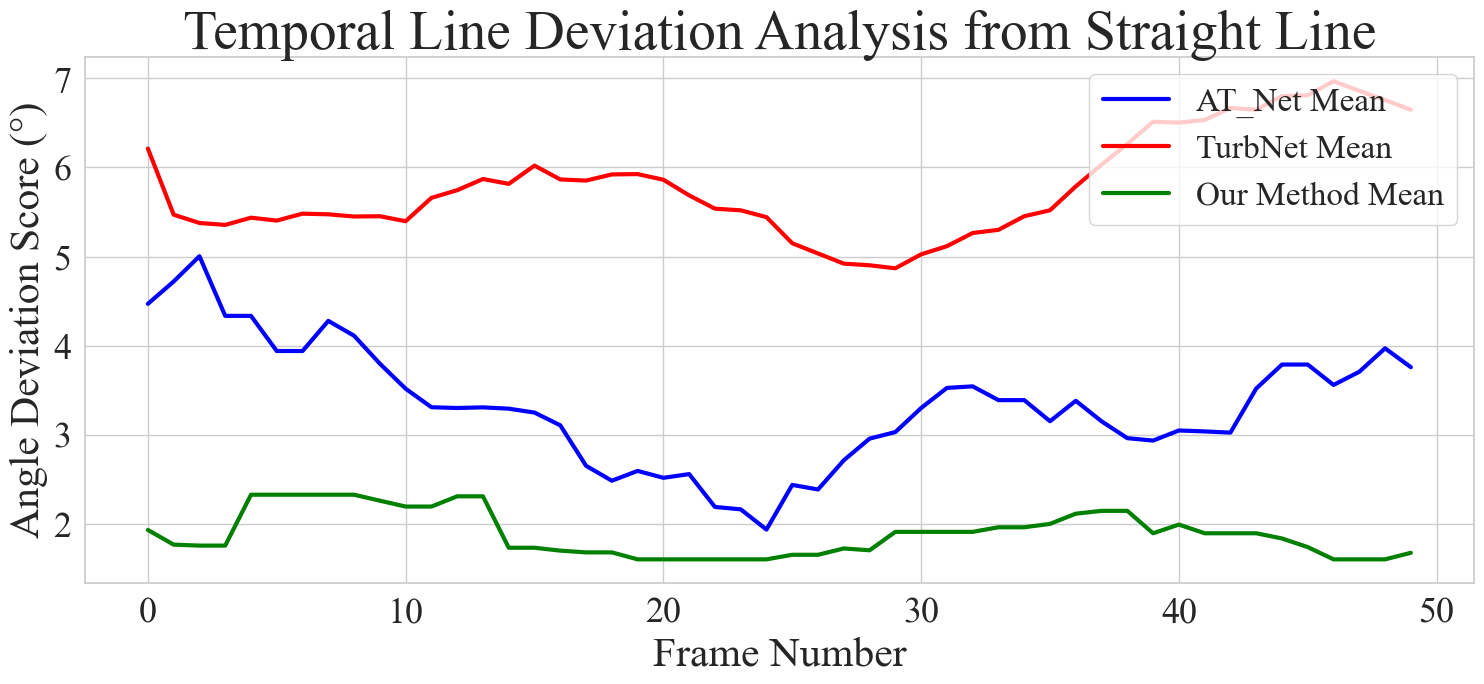}
        \caption{Probabilistic Hough Line Transform-based line deviation metric comparison. Our method exhibits minimal image distortion, particularly in maintaining straight lines, leading to a significantly lower line deviation score compared to competitors.}

    \label{fig:huoghTransformPerfectness}
\end{figure}

\textbf{Line Deviation Metric Comparison:}
~\cref{fig:huoghTransformPerfectness} reveals that for a single image sequence, our method attains a significantly lower average line deviation score than Turbnet and AT-Net. This quantitative metric, complemented by qualitative analysis in ~\cref{fig:GrandPicture}, confirms the superiority of our approach in preserving straight line integrity across frames. Specifically, we report a deviation score of Our: $1.89 (\pm 0.39)$, outperforming AT-Net: $3.33 (\pm 0.33)$ and Turbnet: $5.79 (\pm 0.20)$. This underscores the enhanced geometric accuracy of our image restoration process.

\subsection{Ablation Study}
\label{sec:ablation}
\begin{figure}[htbp]
    \centering
    \includegraphics[width=1\columnwidth]{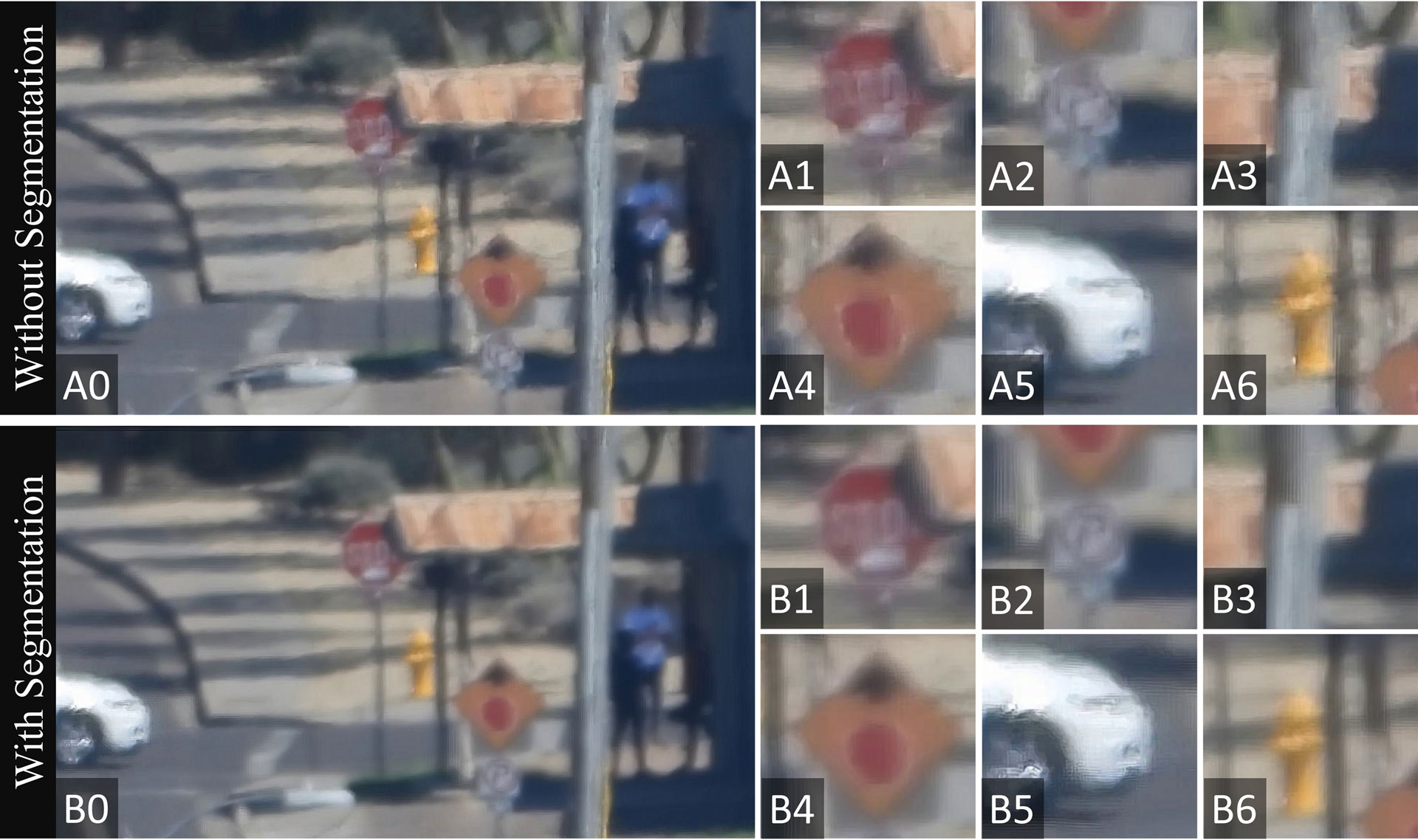}
    \caption{Comparative analysis of image restoration methods in atmospheric turbulence conditions. Panel (A) displays the outcome of a single-frame approach without segmentation, resulting in noticeable artifacts. Panel (B) exhibits the efficacy of our segmentation-based restoration pipeline, significantly enhancing the fidelity of the stop sign with minimal distortions.}
    \label{fig:wwoSegmentation_restormer}
\end{figure}

\textbf{Importance of Segmentation:} In ~\cref{fig:wwoSegmentation_restormer}, we highlight the relative importance of performing segmentation + background enhancement prior to restoration with our trained transformer. The pipeline with segmentation results in sharper background reconstructions on the static STOP sign while still maintaining fidelity for the moving car. 
Choosing the optimal segmentation algorithm for our pipeline raises a crucial question. Despite the simplicity of our AOF method, an ablation study demonstrates its competitiveness with top unsupervised segmentation models in turbulence. In ~\cref{tab:mean_iou}, we compare it against RGA~\cite{dehaoUnsupervised2023}, TMO~\cite{cho2023treating}, Deformable Sprites~\cite{ye2022deformable}, and DS-Net~\cite{LIU2022103700}, where it ranks second in performance but boasts significantly lower latency than the leading RGA method~\cite{dehaoUnsupervised2023}. Note that creating precise ground truth masks is very challenging for human annotators due to the inherent fuzziness of edges in turbulence-distorted images, leading to potential nuisance factors during annotation.

\begin{table}[htbp]
\centering
\caption{Mean IOU Scores for Different Segmentation methods on the URG-T dataset~\cite{dehaoUnsupervised2023}.}
\label{tab:mean_iou}
\resizebox{\columnwidth}{!}{%
\begin{tabular}{|c|c|c|c|c|c|}
\hline
Video Name & RGA~\cite{dehaoUnsupervised2023} & TMO~\cite{cho2023treating} & DSNet~\cite{ye2022deformable} & DSprites~\cite{LIU2022103700} & Ours \\
\hline
Mean IoU & 0.696 & 0.468 & 0.277 & 0.334 & 0.627 \\
\hline
\end{tabular}%
}
\end{table}

\textbf{Importance of Stabilization:} When applied across the entire CLEAR dataset~\cite{anantrasirichai2013atmospheric}, stabilization markedly improved our network's performance. The PSNR value rose from 26.80 to 28.09, and SSIM increased from 0.888 to 0.935, demonstrating stabilization's significant role in enhancing overall image quality.

\textbf{Additional Results:} The supplement provides ablation studies on optical flow choice, visualization of line deviation metrics, a comparison with the P2S simulator~\cite{Simulator0_chimitt2020simulating}, qualitative video comparisons, analysis of restoration network choices, and the impact of stabilization and segmentation.

\section{Conclusion}
This paper presents the first segment-then-restore pipeline for dynamic videos with atmospheric turbulence to obtain enhanced visual reconstruction including the mitigation of geometric distortion as well as recovering high frequency sharpness. In addition to showing the value of segmentation for this problem, our paper introduces a novel tilt-and-blur simulator based on procedural noise (Simplex and Perlin) which enables large scale video dataset creation for training. Compared to state-of-the-art dynamic reconstruction methods, our pipeline is able to synthesize higher quality reconstructions with relatively low latency ($\approx$ 1-2 minutes per video) for $1080p$ resolution video. We release open-source code, simulator, and data available at this link: \href{https://riponcs.github.io/TurbSegRes/}{riponcs.github.io/TurbSegRes}.

We acknowledge potential negative societal impacts, such as privacy concerns arising from enhanced long-range surveillance. We did conduct our research with an ASU IRB STUDY00018456 human subject research exemption. 

Future work will explore enhancing segmentation accuracy with more annotated data, closing the gap between noise-based and physics-based simulation training, and developing real-time restoration for applications in robotics and embedded systems.


\section*{Acknowledgements}
We wish to thank ASU Research Computing for providing GPU resources to support this research. This research is based upon work supported in part by the Office of the Director of National Intelligence (ODNI), Intelligence Advanced Research Projects Activity (IARPA), via 2022-21102100003 and NSF IIS-2232298/2232299/2232300. The views and conclusions contained herein are those of the authors and should not be interpreted as necessarily representing the official policies, either expressed or implied, of ODNI, IARPA, NSF or the U.S. Government. The U.S. Government is authorized to reproduce and distribute reprints for governmental purposes notwithstanding any copyright annotation therein. 


{
    \small
    \bibliographystyle{ieeenat_fullname}
    \bibliography{main}

\begin{thebibliography}{61}
\providecommand{\natexlab}[1]{#1}
\providecommand{\url}[1]{\texttt{#1}}
\expandafter\ifx\csname urlstyle\endcsname\relax
  \providecommand{\doi}[1]{doi: #1}\else
  \providecommand{\doi}{doi: \begingroup \urlstyle{rm}\Url}\fi

\bibitem[Anantrasirichai(2023)]{ANANTRASIRICHAI202369}
Nantheera Anantrasirichai.
\newblock Atmospheric turbulence removal with complex-valued convolutional neural network.
\newblock \emph{Pattern Recognition Letters}, 171:\penalty0 69--75, 2023.

\bibitem[Anantrasirichai et~al.(2013)Anantrasirichai, Achim, Kingsbury, and Bull]{anantrasirichai2013atmospheric}
Nantheera Anantrasirichai, Alin Achim, Nick~G Kingsbury, and David~R Bull.
\newblock Atmospheric turbulence mitigation using complex wavelet-based fusion.
\newblock \emph{IEEE Transactions on Image Processing}, 22\penalty0 (6):\penalty0 2398--2408, 2013.

\bibitem[Aubailly et~al.(2009)Aubailly, Vorontsov, Carhart, and Valley]{aubailly2009automated}
Mathieu Aubailly, Mikhail~A Vorontsov, Gary~W Carhart, and Michael~T Valley.
\newblock Automated video enhancement from a stream of atmospherically-distorted images: the lucky-region fusion approach.
\newblock \emph{Atmospheric Optics: Models, Measurements, and Target-in-the-Loop Propagation III}, 7463:\penalty0 104--113, 2009.

\bibitem[Bensimon et~al.(1981)Bensimon, Englander, Karoubi, and Weiss]{bensimon1981measurement}
David Bensimon, A Englander, R Karoubi, and Meira Weiss.
\newblock Measurement of the probability of getting a lucky short-exposure image through turbulence.
\newblock \emph{JOSA}, 71\penalty0 (9):\penalty0 1138--1139, 1981.

\bibitem[Bos and Roggemann(2012)]{Simulator4_bos2012technique}
Jeremy~P Bos and Michael~C Roggemann.
\newblock Technique for simulating anisoplanatic image formation over long horizontal paths.
\newblock \emph{Optical Engineering}, 51\penalty0 (10):\penalty0 101704--101704, 2012.

\bibitem[Chan(2022)]{chan2022tilt}
Stanley~H Chan.
\newblock Tilt-then-blur or blur-then-tilt? clarifying the atmospheric turbulence model.
\newblock \emph{IEEE Signal Processing Letters}, 29:\penalty0 1833--1837, 2022.

\bibitem[Chen et~al.(2014)Chen, Haik, and Yitzhaky]{chen2014detecting}
Eli Chen, Oren Haik, and Yitzhak Yitzhaky.
\newblock Detecting and tracking moving objects in long-distance imaging through turbulent medium.
\newblock \emph{Applied Optics}, 53\penalty0 (6):\penalty0 1181--1190, 2014.

\bibitem[Chimitt and Chan(2020)]{Simulator0_chimitt2020simulating}
Nicholas Chimitt and Stanley~H Chan.
\newblock Simulating anisoplanatic turbulence by sampling intermodal and spatially correlated zernike coefficients.
\newblock \emph{Optical Engineering}, 59\penalty0 (8):\penalty0 083101, 2020.

\bibitem[Chimitt et~al.(2022)Chimitt, Zhang, Mao, and Chan]{9969142}
Nicholas Chimitt, Xingguang Zhang, Zhiyuan Mao, and Stanley~H. Chan.
\newblock Real-time dense field phase-to-space simulation of imaging through atmospheric turbulence.
\newblock \emph{IEEE Transactions on Computational Imaging}, 8:\penalty0 1159--1169, 2022.

\bibitem[Cho et~al.(2023)Cho, Lee, Lee, Park, Kim, and Lee]{cho2023treating}
Suhwan Cho, Minhyeok Lee, Seunghoon Lee, Chaewon Park, Donghyeong Kim, and Sangyoun Lee.
\newblock Treating motion as option to reduce motion dependency in unsupervised video object segmentation.
\newblock In \emph{Proceedings of the IEEE/CVF Winter Conference on Applications of Computer Vision}, pages 5140--5149, 2023.

\bibitem[Crittenden~Jr et~al.(1978)Crittenden~Jr, Cooper, Rodeback, Kalmbach, and Armstead]{crittenden1978effects}
EC Crittenden~Jr, AW Cooper, GW Rodeback, SH Kalmbach, and RL Armstead.
\newblock Effects of turbulence on imaging through the atmosphere.
\newblock In \emph{Optical Properties of the Atmosphere}, pages 130--134. SPIE, 1978.

\bibitem[Cui and Zhang(2019)]{8901108}
Linyan Cui and Yan Zhang.
\newblock Accurate semantic segmentation in turbulence media.
\newblock \emph{IEEE Access}, 7:\penalty0 166749--166761, 2019.

\bibitem[Fazlali et~al.(2022)Fazlali, Shirani, Bradford, and Kirubarajan]{MultiDeep2022}
Hamidreza Fazlali, Shahram Shirani, Michael Bradford, and Thia Kirubarajan.
\newblock Atmospheric turbulence removal in long-range imaging using a data-driven-based approach.
\newblock \emph{International Journal of Computer Vision}, 130\penalty0 (4):\penalty0 1031--1049, 2022.

\bibitem[Fried(1965)]{fried1965statistics}
David~L Fried.
\newblock Statistics of a geometric representation of wavefront distortion.
\newblock \emph{JOSA}, 55\penalty0 (11):\penalty0 1427--1435, 1965.

\bibitem[Fried(1978)]{fried1978probability}
David~L Fried.
\newblock Probability of getting a lucky short-exposure image through turbulence.
\newblock \emph{JOSA}, 68\penalty0 (12):\penalty0 1651--1658, 1978.

\bibitem[Fried(1982)]{fried1982anisoplanatism}
David~L Fried.
\newblock Anisoplanatism in adaptive optics.
\newblock \emph{JOSA}, 72\penalty0 (1):\penalty0 52--61, 1982.

\bibitem[Furhad et~al.(2016)Furhad, Tahtali, and Lambert]{MultiKmeanBspine2016}
Md~Hasan Furhad, Murat Tahtali, and Andrew Lambert.
\newblock Restoring atmospheric-turbulence-degraded images.
\newblock \emph{Applied Optics}, 55\penalty0 (19):\penalty0 5082--5090, 2016.

\bibitem[Gal et~al.(2014)Gal, Kiryati, and Sochen]{MultiRegsiDeblur2014}
Ronen Gal, Nahum Kiryati, and Nir Sochen.
\newblock Progress in the restoration of image sequences degraded by atmospheric turbulence.
\newblock \emph{Pattern Recognition Letters}, 48:\penalty0 8--14, 2014.

\bibitem[Gilles and Ferrante(2017)]{OTIS2017}
J{\'e}r{\^o}me Gilles and Nicholas~B Ferrante.
\newblock Open turbulent image set (otis).
\newblock \emph{Pattern Recognition Letters}, 86:\penalty0 38--41, 2017.

\bibitem[Gray(2019)]{Simulator10_2019zernikecalc}
R Gray.
\newblock Zernikecalc: a matlab function to work with zernike polynomials over circular and non-circular pupils, 2019.

\bibitem[Gutierrez et~al.(2006)Gutierrez, Seron, Munoz, and Anson]{gutierrez2006simulation}
Diego Gutierrez, Francisco~J Seron, Adolfo Munoz, and Oscar Anson.
\newblock Simulation of atmospheric phenomena.
\newblock \emph{Computers \& Graphics}, 30\penalty0 (6):\penalty0 994--1010, 2006.

\bibitem[Hardie et~al.(2017)Hardie, Power, LeMaster, Droege, Gladysz, and Bose-Pillai]{Simulator6_hardie2017simulation}
Russell~C Hardie, Jonathan~D Power, Daniel~A LeMaster, Douglas~R Droege, Szymon Gladysz, and Santasri Bose-Pillai.
\newblock Simulation of anisoplanatic imaging through optical turbulence using numerical wave propagation with new validation analysis.
\newblock \emph{Optical Engineering}, 56\penalty0 (7):\penalty0 071502--071502, 2017.

\bibitem[Hunt et~al.(2018)Hunt, Iler, Bailey, and Rucci]{Simulator9_2018synthesis}
Bobby~R Hunt, Amber~L Iler, Christopher~A Bailey, and Michael~A Rucci.
\newblock Synthesis of atmospheric turbulence point spread functions by sparse and redundant representations.
\newblock \emph{Optical Engineering}, 57\penalty0 (2):\penalty0 024101, 2018.

\bibitem[Ihrke et~al.(2007)Ihrke, Ziegler, Tevs, Theobalt, Magnor, and Seidel]{ihrke2007eikonal}
Ivo Ihrke, Gernot Ziegler, Art Tevs, Christian Theobalt, Marcus Magnor, and Hans-Peter Seidel.
\newblock Eikonal rendering: Efficient light transport in refractive objects.
\newblock \emph{ACM Transactions on Graphics (TOG)}, 26\penalty0 (3):\penalty0 59--es, 2007.

\bibitem[Ishimaru(1999)]{ishimaru1999wave}
A. Ishimaru.
\newblock \emph{Wave Propagation and Scattering in Random Media}.
\newblock Wiley, 1999.

\bibitem[Jiang et~al.(2023)Jiang, Boominathan, and Veeraraghavan]{jiang2023nert}
Weiyun Jiang, Vivek Boominathan, and Ashok Veeraraghavan.
\newblock Nert: Implicit neural representations for unsupervised atmospheric turbulence mitigation.
\newblock In \emph{Proceedings of the IEEE/CVF Conference on Computer Vision and Pattern Recognition}, pages 4235--4242, 2023.

\bibitem[{Kolmogorov}(1941)]{kolmogorov}
Andrey~Nikolaevich {Kolmogorov}.
\newblock {Dissipation of Energy in Locally Isotropic Turbulence}.
\newblock \emph{Akademiia Nauk SSSR Doklady}, 32:\penalty0 16, 1941.

\bibitem[Kolmogorov(1991)]{kolmogorov1991local}
Andrei~Nikolaevich Kolmogorov.
\newblock The local structure of turbulence in incompressible viscous fluid for very large reynolds numbers.
\newblock \emph{Proceedings of the Royal Society of London. Series A: Mathematical and Physical Sciences}, 434\penalty0 (1890):\penalty0 9--13, 1991.

\bibitem[Kopeika(1998)]{kopeika1998system}
N.S. Kopeika.
\newblock \emph{A System Engineering Approach to Imaging}.
\newblock SPIE Optical Engineering Press, 1998.

\bibitem[Lagae et~al.(2010)Lagae, Lefebvre, Cook, DeRose, Drettakis, Ebert, Lewis, Perlin, and Zwicker]{lagae2010survey}
Ares Lagae, Sylvain Lefebvre, Rob Cook, Tony DeRose, George Drettakis, David~S Ebert, John~P Lewis, Ken Perlin, and Matthias Zwicker.
\newblock A survey of procedural noise functions.
\newblock \emph{Computer Graphics Forum}, 29\penalty0 (8):\penalty0 2579--2600, 2010.

\bibitem[Lau et~al.(2019)Lau, Lai, and Lui]{MultiRPCAStabilization2019}
Chun~Pong Lau, Yu~Hin Lai, and Lok~Ming Lui.
\newblock Restoration of atmospheric turbulence-distorted images via rpca and quasiconformal maps.
\newblock \emph{Inverse Problems}, 35\penalty0 (7):\penalty0 074002, 2019.

\bibitem[Leonard et~al.(2012)Leonard, Howe, and Oxford]{Simulator5_leonard2012simulation}
Kevin~R Leonard, Jonathan Howe, and David~E Oxford.
\newblock Simulation of atmospheric turbulence effects and mitigation algorithms on stand-off automatic facial recognition.
\newblock In \emph{Optics and Photonics for Counterterrorism, Crime Fighting, and Defence VIII}, pages 182--198. SPIE, 2012.

\bibitem[Li et~al.(2007)Li, Mersereau, and Simske]{MultiPCA2007atmospheric}
Dalong Li, Russell~M Mersereau, and Steven Simske.
\newblock Atmospheric turbulence-degraded image restoration using principal components analysis.
\newblock \emph{IEEE Geoscience and Remote Sensing Letters}, 4\penalty0 (3):\penalty0 340--344, 2007.

\bibitem[Li et~al.(2021)Li, Thapa, Whyte, Reed, Jayasuriya, and Ye]{Li_2021_ICCV}
Nianyi Li, Simron Thapa, Cameron Whyte, Albert~W. Reed, Suren Jayasuriya, and Jinwei Ye.
\newblock Unsupervised non-rigid image distortion removal via grid deformation.
\newblock In \emph{Proceedings of the IEEE/CVF International Conference on Computer Vision (ICCV)}, pages 2522--2532, 2021.

\bibitem[Liu et~al.(2022)Liu, Wang, Wang, and Su]{LIU2022103700}
Jing Liu, Jiaxiang Wang, Weikang Wang, and Yuting Su.
\newblock Ds-net: Dynamic spatiotemporal network for video salient object detection.
\newblock \emph{Digital Signal Processing}, 130:\penalty0 103700, 2022.

\bibitem[Mao et~al.(2020)Mao, Chimitt, and Chan]{TCIMultiPurdue2020}
Zhiyuan Mao, Nicholas Chimitt, and Stanley~H Chan.
\newblock Image reconstruction of static and dynamic scenes through anisoplanatic turbulence.
\newblock \emph{IEEE Transactions on Computational Imaging}, 6:\penalty0 1415--1428, 2020.

\bibitem[Mao et~al.(2021)Mao, Chimitt, and Chan]{mao2021accelerating}
Zhiyuan Mao, Nicholas Chimitt, and Stanley~H Chan.
\newblock Accelerating atmospheric turbulence simulation via learned phase-to-space transform.
\newblock In \emph{Proceedings of the IEEE/CVF International Conference on Computer Vision}, pages 14759--14768, 2021.

\bibitem[Mao et~al.(2022)Mao, Jaiswal, Wang, and Chan]{Turbnet2022}
Zhiyuan Mao, Ajay Jaiswal, Zhangyang Wang, and Stanley~H Chan.
\newblock Single frame atmospheric turbulence mitigation: A benchmark study and a new physics-inspired transformer model.
\newblock In \emph{European Conference on Computer Vision}, pages 430--446. Springer, 2022.

\bibitem[Matas et~al.(2000)Matas, Galambos, and Kittler]{Matas2000RobustDO}
Jiri Matas, Charles Galambos, and Josef Kittler.
\newblock Robust detection of lines using the progressive probabilistic hough transform.
\newblock \emph{Comput. Vis. Image Underst.}, 78:\penalty0 119--137, 2000.

\bibitem[Olano et~al.(2002)Olano, Hart, Heidrich, Mark, and Perlin]{olano2002real}
Mark Olano, John~C Hart, Wolfgang Heidrich, Bill Mark, and Ken Perlin.
\newblock Real-time shading languages.
\newblock \emph{SIGGRAPH Course Notes}, 2002.

\bibitem[P{\'e}rez et~al.(2023)P{\'e}rez, Gangnet, and Blake]{blending2023poisson}
Patrick P{\'e}rez, Michel Gangnet, and Andrew Blake.
\newblock Poisson image editing.
\newblock In \emph{Seminal Graphics Papers: Pushing the Boundaries, Volume 2}, pages 577--582. 2023.

\bibitem[Perlin(1985)]{PerlinNoise}
Ken Perlin.
\newblock An image synthesizer.
\newblock In \emph{Proceedings of the 12th Annual Conference on Computer Graphics and Interactive Techniques}, page 287–296, New York, NY, USA, 1985. Association for Computing Machinery.

\bibitem[Potvin et~al.(2007)Potvin, Forand, and Dion]{potvin2007parametric}
G Potvin, JL Forand, and D Dion.
\newblock A parametric model for simulating turbulence effects on imaging systems.
\newblock \emph{DRDC Valcartier TR 2006}, 787, 2007.

\bibitem[Potvin et~al.(2011)Potvin, Forand, and Dion]{Simulator2_2011Simple}
Guy Potvin, J~Luc Forand, and Denis Dion.
\newblock A simple physical model for simulating turbulent imaging.
\newblock In \emph{Infrared Imaging Systems: Design, Analysis, Modeling, and Testing XXII}, pages 323--335. SPIE, 2011.

\bibitem[Qin et~al.(2023)Qin, Saha, Jayasuriya, Ye, and Li]{dehaoUnsupervised2023}
Dehao Qin, Ripon Saha, Suren Jayasuriya, Jinwei Ye, and Nianyi Li.
\newblock Unsupervised region-growing network for object segmentation in atmospheric turbulence.
\newblock \emph{arXiv:2311.03572}, 2023.

\bibitem[Reinhardt et~al.(2016)Reinhardt, Hammel, and Tsintikidis]{reinhardt2016efficient}
Colin~N Reinhardt, Stephen~M Hammel, and Dimitris Tsintikidis.
\newblock Efficient physics-based predictive 3d image modeling and simulation of optical atmospheric refraction phenomena.
\newblock In \emph{Laser Communication and Propagation through the Atmosphere and Oceans V}, page 99790S. International Society for Optics and Photonics, 2016.

\bibitem[Repasi and Weiss(2011)]{repasi2011computer}
Endre Repasi and Robert Weiss.
\newblock Computer simulation of image degradations by atmospheric turbulence for horizontal views.
\newblock In \emph{Infrared Imaging Systems: Design, Analysis, Modeling, and Testing XXII}, page 80140U. International Society for Optics and Photonics, 2011.

\bibitem[Roberto~e Souza et~al.(2022)Roberto~e Souza, Maia, and Pedrini]{videostabilzationsurvey}
Marcos Roberto~e Souza, Helena de~Almeida Maia, and Helio Pedrini.
\newblock Survey on digital video stabilization: Concepts, methods, and challenges.
\newblock \emph{ACM Comput. Surv.}, 55\penalty0 (3), 2022.

\bibitem[Roggemann et~al.(1996)Roggemann, Welsh, and Hunt]{roggemann1996imaging}
M.C. Roggemann, B.M. Welsh, and B.R. Hunt.
\newblock \emph{Imaging Through Turbulence}.
\newblock Taylor \& Francis, 1996.

\bibitem[Roggemann(1992)]{Simulator1_2011Optical}
Michael~C Roggemann.
\newblock Optical performance of fully and partially compensated adaptive optics systems using least-squares and minimum variance phase reconstructors.
\newblock \emph{Computers \& Electrical Engineering}, 18\penalty0 (6):\penalty0 451--466, 1992.

\bibitem[Saha et~al.(2022)Saha, Salcin, Kim, Smith, and Jayasuriya]{saha2022turbulence}
Ripon~Kumar Saha, Esen Salcin, Jihoo Kim, Joseph Smith, and Suren Jayasuriya.
\newblock Turbulence strength $c_n^2$ estimation from video using physics-based deep learning.
\newblock \emph{Optics Express}, 30\penalty0 (22):\penalty0 40854--40870, 2022.

\bibitem[Schwartzman et~al.(2017)Schwartzman, Alterman, Zamir, and Schechner]{schwartzman2017turbulence}
Armin Schwartzman, Marina Alterman, Rotem Zamir, and Yoav~Y Schechner.
\newblock Turbulence-induced 2d correlated image distortion.
\newblock In \emph{2017 IEEE International Conference on Computational Photography (ICCP)}, pages 1--13. IEEE, 2017.

\bibitem[Tatarski(2016)]{tatarski2016wave}
Valerian~Ilich Tatarski.
\newblock \emph{Wave propagation in a turbulent medium}.
\newblock Courier Dover Publications, 2016.

\bibitem[Teed and Deng(2020)]{teed2020raft}
Zachary Teed and Jia Deng.
\newblock Raft: Recurrent all-pairs field transforms for optical flow.
\newblock In \emph{European conference on computer vision}, pages 402--419. Springer, 2020.

\bibitem[Xie et~al.(2016)Xie, Zhang, Tao, Hu, Qu, and Wang]{xie2016removing}
Yuan Xie, Wensheng Zhang, Dacheng Tao, Wenrui Hu, Yanyun Qu, and Hanzi Wang.
\newblock Removing turbulence effect via hybrid total variation and deformation-guided kernel regression.
\newblock \emph{IEEE Transactions on Image Processing}, 25\penalty0 (10):\penalty0 4943--4958, 2016.

\bibitem[Yasarla and Patel(2021)]{AtNet_2021learning}
Rajeev Yasarla and Vishal~M Patel.
\newblock Learning to restore images degraded by atmospheric turbulence using uncertainty.
\newblock In \emph{2021 IEEE International Conference on Image Processing (ICIP)}, pages 1694--1698. IEEE, 2021.

\bibitem[Ye et~al.(2022)Ye, Li, Tucker, Kanazawa, and Snavely]{ye2022deformable}
Vickie Ye, Zhengqi Li, Richard Tucker, Angjoo Kanazawa, and Noah Snavely.
\newblock Deformable sprites for unsupervised video decomposition.
\newblock In \emph{Proceedings of the IEEE/CVF Conference on Computer Vision and Pattern Recognition}, pages 2657--2666, 2022.

\bibitem[Zamek and Yitzhaky(2006)]{zamek2006turbulence}
Steve Zamek and Yitzhak Yitzhaky.
\newblock Turbulence strength estimation from an arbitrary set of atmospherically degraded images.
\newblock \emph{JOSA A}, 23\penalty0 (12):\penalty0 3106--3113, 2006.

\bibitem[Zamir et~al.(2022)Zamir, Arora, Khan, Hayat, Khan, and Yang]{zamir2022restormer}
Syed~Waqas Zamir, Aditya Arora, Salman Khan, Munawar Hayat, Fahad~Shahbaz Khan, and Ming-Hsuan Yang.
\newblock Restormer: Efficient transformer for high-resolution image restoration.
\newblock In \emph{Proceedings of the IEEE/CVF Conference on Computer Vision and Pattern Recognition}, pages 5728--5739, 2022.

\bibitem[Zhang et~al.(2024)Zhang, Mao, Chimitt, and Chan]{zhang2024imaging}
Xingguang Zhang, Zhiyuan Mao, Nicholas Chimitt, and Stanley~H Chan.
\newblock Imaging through the atmosphere using turbulence mitigation transformer.
\newblock \emph{IEEE Transactions on Computational Imaging}, 2024.

\bibitem[Zhu and Milanfar(2012)]{zhu2012removing}
Xiang Zhu and Peyman Milanfar.
\newblock Removing atmospheric turbulence via space-invariant deconvolution.
\newblock \emph{IEEE Transactions on Pattern Analysis and Machine Intelligence}, 35\penalty0 (1):\penalty0 157--170, 2012.

\end{thebibliography}
}


\end{document}